%% file: main.tex
\documentclass[10pt,twocolumn,letterpaper]{article}

\usepackage{iccv}      

\input{preamble}

\definecolor{iccvblue}{rgb}{0.21,0.49,0.74}
\usepackage[pagebackref,breaklinks,colorlinks,allcolors=iccvblue]{hyperref}

\title{Deformable Mamba for Wide Field of View Segmentation}

\author{Jie Hu$^{1,*}$, 
~~Junwei Zheng$^{1,*}$
~~Jiale Wei$^1$,
~~Jiaming Zhang$^{1,2,\dag}$
~~Rainer Stiefelhagen$^1$\\
\normalsize
$^1$Karlsruhe Institute of Technology,
\normalsize
$^2$ETH Zurich\\
\url{https://github.com/JieHu1996/DeformableMamba}
}

\begin{document}

\maketitle

{
  \renewcommand{\thefootnote}
    {\fnsymbol{footnote}}
  \footnotetext[1]{Equal contribution.}
  \footnotetext[2]{Corresponding author (e-mail: {\tt jiaming.zhang@kit.edu}).}
}

\input{sec/0_abstract} 

\begin{figure}[t]
    \centering
    \begin{subfigure}{\linewidth}
        \centering
        \includegraphics[width=\linewidth]{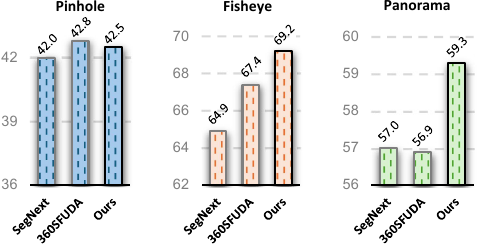}
        \caption{Comparison across narrow- and wide-FoV semantic segmentation.}
        \label{fig1a}
    \end{subfigure}

    \begin{subfigure}{\linewidth}
        \centering
        \includegraphics[width=\linewidth]{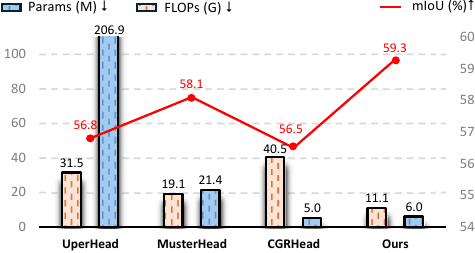}  
        \caption{Performance and efficiency comparison of different decoders on Stanford2D3D~\cite{armeni2017joint} panoramic dataset with the same backbone~\cite{liu2024vmamba}.}
        \label{fig1b}
    \end{subfigure}
    
    \caption{
    Our deformable Mamba (a) achieves better results across wide-FoV semantic segmentation while (b) maintaining parameter and computational efficiency. 
    }
    \label{fig1}
    \vskip -4ex
\end{figure}

\input{sec/1_intro}

\begin{figure*}[!t]
    \centering
    \includegraphics[width=1.0\textwidth]{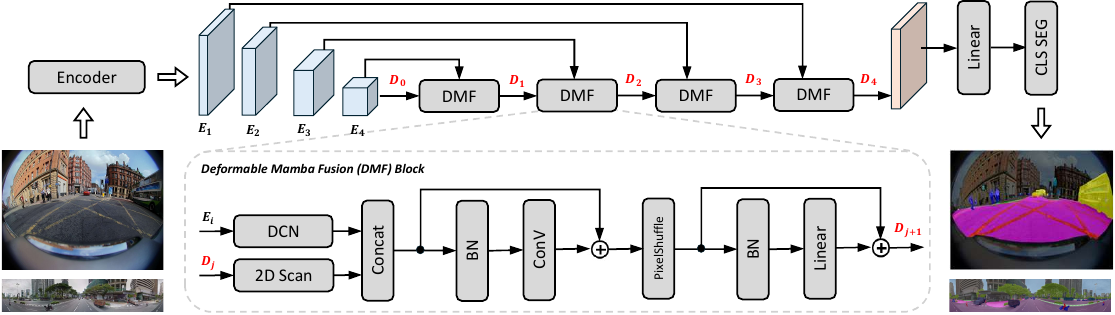}
    \vskip -2ex
    \caption{Overview of the Deformable Mamba (DMamba) framework. Given wide-FoV images (180{\textdegree} or 360{\textdegree}), the features extracted by an encoder, are fused by the proposed Deformable Mamba Decoder, which constructed by four Deformable Mamba Fusion (DMF) modules.} 
    \label{fig3}

\end{figure*}

\input{sec/2_related_work}
\input{sec/3_methodology}
\input{sec/4_experiments}
\input{sec/5_conclusion}

{
    \small
    \bibliographystyle{ieeenat_fullname}
    \bibliography{main}
}

\end{document}

%% file: preamble.tex
\usepackage{graphicx}
\usepackage{amsmath}
\usepackage{amssymb}
\usepackage{booktabs}
\usepackage{mathtools}
\usepackage[accsupp]{axessibility}  
\usepackage{color}
\usepackage{colortbl}
\usepackage{makecell} 
\usepackage{array}
\usepackage{pdflscape}
\usepackage[longtable]{multirow}
\usepackage{longtable}
\usepackage{tabularray}
\usepackage{booktabs}
\usepackage{times}
\usepackage{epsfig}
\usepackage{amsmath}
\usepackage{bbm}
\DeclareMathAlphabet\mathbfcal{OMS}{cmsy}{b}{n}
\usepackage{makecell}
\usepackage{siunitx}
\usepackage{svg}
\usepackage{float}
\usepackage{comment}
\usepackage{lipsum}
\usepackage{stfloats}
\usepackage{multirow}
\usepackage{bm}
\usepackage{etoolbox}
\usepackage{icomma}
\usepackage{array}
\usepackage{tabulary}
\usepackage{xcolor}
\usepackage{paralist}
\usepackage{booktabs}
\usepackage{adjustbox}
\usepackage{caption}
\usepackage{subcaption}
\usepackage{arydshln}
\usepackage{rotating}

\definecolor{gray}{rgb}{0.3,0.3,0.3}
\definecolor{blue}{rgb}{0,0.5,1}
\definecolor{mask_red}{rgb}{1,0,0.8}
\definecolor{green}{rgb}{0.2,1,0.2}
\definecolor{rblue}{rgb}{0,0,1}
\definecolor{lightblue}{HTML}{6495ed}
\definecolor{lightred}{HTML}{F19C99}

\definecolor{graytablerow}{gray}{0.6}

\usepackage{pifont}
\usepackage{tabu}
\usepackage[pro]{fontawesome5}

\usepackage{tikz}

%% file: sec/0_abstract.tex
\begin{abstract}\label{sec:abstract}
Recent advancements in the Mamba architecture, with its linear computational complexity, being a promising alternative to transformer architectures suffering from quadratic complexity. While existing works primarily focus on adapting Mamba as vision encoders, the critical role of task-specific Mamba decoders remains under-explored, particularly for distortion-prone dense prediction tasks. This paper addresses two interconnected challenges: (1) The design of a Mamba-based decoder that seamlessly adapts to various architectures (e.g., CNN-, Transformer-, and Mamba-based backbones), and (2) The performance degradation in decoders lacking distortion-aware capability when processing wide-FoV images (e.g., 180{\textdegree} fisheye and 360{\textdegree} panoramic settings). We propose the Deformable Mamba Decoder, an efficient distortion-aware decoder that integrates Mamba's computational efficiency with adaptive distortion awareness. Comprehensive experiments on five wide-FoV segmentation benchmarks validate its effectiveness. Notably, our decoder achieves a \textbf{{+}2.5\%} performance improvement on the 360{\textdegree} Stanford2D3D segmentation benchmark while reducing \textbf{72\%} parameters and \textbf{97\%} FLOPs, as compared to the widely-used decoder heads.

\end{abstract}

%% file: sec/1_intro.tex
\section{Introduction}
\label{sec:intro}

Dense image analysis like image segmentation~\cite{badrinarayanan2017segnet,xie2021segformer,guo2022segnext} is a fundamental task in computer vision, forming a crucial component of numerous downstream vision-based applications~\cite{zheng2024materobot}. At the same time, advancements in sensor technology have led to a variety of sensor types, each with unique characteristics that impact imaging. In recent years, diverse sensors such as pinhole cameras, fisheye cameras~\cite{yogamani2019woodscape, sekkat2022synwoodscape} with 180{\textdegree} field of views (FoV), and even panoramic cameras~\cite{chang2017matterport3d, armeni2017joint, zhang2024behind} with 360{\textdegree} FoV have been increasingly adopted across vision tasks. 

\begin{figure}[t]
  \centering
  \includegraphics[width=\linewidth]{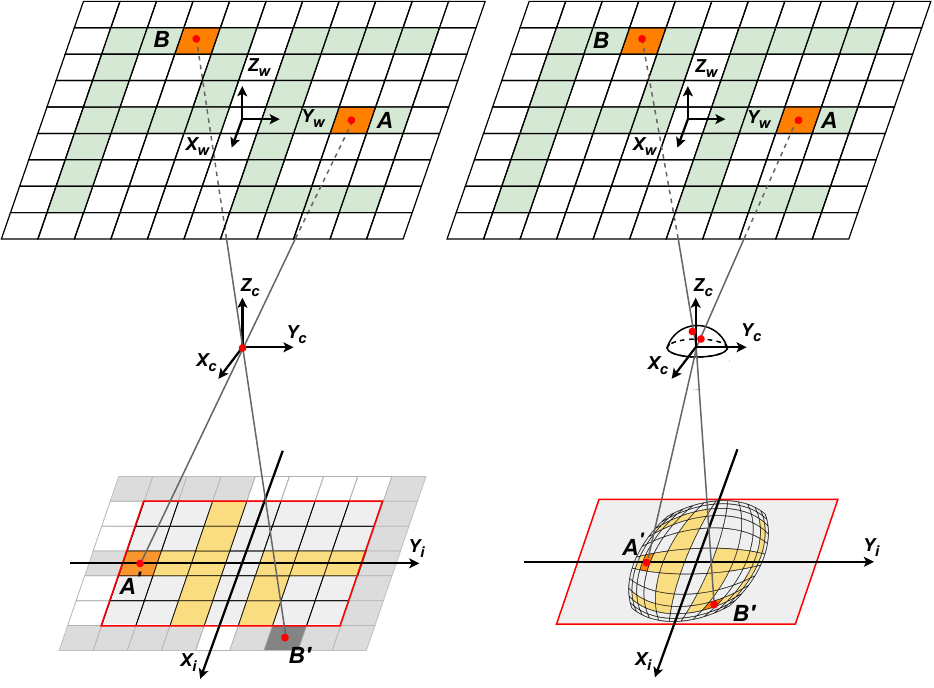}
  \vskip -0.5em
  \caption{Comparison of camera imaging with narrow-FoV (left) and wide-FoV (right) cameras. Narrow-FoV cameras maintain geometric fidelity but limited coverage, whereas wide-FoV cameras offer expansive scene capture while introducing substantial geometric distortions.}
  \vskip -1.5em
  \label{fig2}
\end{figure}

Based on our experiments (Fig.~\ref{fig1a}), we observe that directly applying models designed for narrow-FoV data to wide-FoV data often leads to a suboptimal solution. For instance, SegNext~\cite{guo2022segnext} specifically designed for narrow-FoV pinhole cameras typically cannot generalize well when used for analyzing fisheye images with 180{\textdegree} FoV or panoramic images with 360{\textdegree} FoV. This is primarily due to the high degree of image distortion and object deformation present in wide-FoV cameras. As shown in Fig.~\ref{fig2}, the altered pixel arrangement in wide-FoV images might lead to misclassifications and limits adaptability of narrow-FoV models~\cite{badrinarayanan2017segnet, chen2018encoder, chen2023cyclemlp, zheng2021rethinking, strudel2021segmenter}. 
To address this, previous methods have spent a large effort in architecture design to fit specific sensor types, such as CNN-based and transformer-based models~\cite{tarvainen2017mean, kumar2021omnidet} for fisheye images, as well as approaches~\cite{zhang2022bending,zhang2024behind} tailored to mitigate distortion in panoramic segmentation. 

However, most existing methods deeply couple distortion-aware capability with the entire model (both encoder and decoder). The drawbacks are two-fold: 
(1) These models are often designed for specific FoVs, yet hard to adapt to varying distortion characteristics. 
For example, narrow-FoV model SegNext~\cite{guo2022segnext} and wide-FoV model 360SFUDA~\cite{zheng2024semantics} cannot consistently perform well across different wide-FoV segmentation (Fig.~\ref{fig1a}). 

(2) The tight coupling with the overall architecture limits the reusability of large-scale pretrained backbones, as any modification requires retraining. Therefore, in this work, we focus on flexible decoder designs and constrain distortion awareness to the decoder level, yielding greater flexibility in utilizing diverse pretrained backbones, striking a tradeoff between adaptability and modularity. 

Furthermore, we mainly utilize the Mamba~\cite{gu2023mamba} mechanism due to its benefit of linear computational complexity. However, most previous works~\cite{vim, liu2024vmamba, yang2024plainmamba, huang2024localmamba, li2024videomamba, chen2024rsmamba, zhu2024samba} have focused on Mamba-based encoders, research on corresponding decoders for downstream tasks remains limited. We contend that decoders are equally crucial as encoders in the context of dense prediction tasks. As illustrated in Fig.~\ref{fig1b}, a widely used decoder UperHead~\cite{xiao2018unified} excels at multi-scale feature fusion but suffers from excessive computational overhead. Alternative methods~\cite{xu2022uperformer, ni2024context} applied lightweight CNN- and transformer-based decoders to reduce computational cost. However, they suffer notable performance degradation when processing wide-FoV data. 

To address these challenges, we propose the \textbf{Deformable Mamba Decoder} , an efficient distortion-aware decoder for image segmentation across varying wide-FoVs. In this work, a unified framework is created to effectively handling images from multiple camera types and sources, spanning FoVs from 180{\textdegree} to 360{\textdegree}, from indoor to outdoor scenarios, in synthetic and real-world imageries. Our deformable decoder can seamlessly integrate with CNN-, transformer-, and Mamba-based backbones, imparting distortion-awareness capability to the overall model.

To evaluate the effectiveness of our method, we conduct extensive experiments across five datasets for wide-FoV segmentation, including two 180{\textdegree} (WoodScape~\cite{yogamani2019woodscape}, SynWoodScape~\cite{sekkat2022synwoodscape}) datasets and three 360{\textdegree} ( Stanford2D3D~\cite{armeni2017joint}, Matterport3D~\cite{chang2017matterport3d}, SynPASS~\cite{zhang2024behind}). Our framework performs consistently well across various datasets. Notably, while reducing the decoder parameters by $72\%$ and FLOPs by $97\%$, Deformable Mamba achieved state-of-the-art performance, with a ${+}2.5\%$ absolute improvement over previous VMamba~\cite{liu2024vmamba} on the 360{\textdegree} Stanford2D3D dataset.  
To summarize, our contributions are as follows. 

\begin{compactitem}
    \item We propose an efficient decoder based on the Mamba architecture, which addresses the gap in decoder-oriented Mamba-based efficient approaches.
    \item For the first time, we incorporate distortion-awareness exclusively at the Mamba decoder level, decoupling it from the overall framework, 
    which enables integration with CNN-, Transformer-, and Mamba-based backbones to enhance distortion perception.
    \item Extensive experiments on five datasets across a variety of indoor and outdoor, synthetic and real-world wide-FoV scenes, demonstrate the state-of-the-art performance of our method in wide-FoV segmentation.
\end{compactitem}

%% file: sec/2_related_work.tex
\section{Related Work}
\label{sec:related_work}

\subsection{Wide Field of View Segmentation} 
Wide-FoV image segmentation has received increasing attention as it enables broader scene understanding in applications such as autonomous driving and robotics. Wide-FoV images, like panoramic and fisheye images, offer expansive visual coverage but introduce unique challenges due to geometric distortions, spatial resolution variations, and uneven feature distribution.
Early segmentation methods~\cite{long2015fully, badrinarayanan2017segnet, chen2018encoder, fu2019dual, huang2019ccnet, yuan2021ocnet, yuan2021ocnet, liu2020covariance, li2021ctnet, zheng2021rethinking, strudel2021segmenter, xie2021segformer, cheng2021per, gu2022multi, chen2021cyclemlp, zheng2024materobot} have achieved significant success on narrow-FoV image segmentation. However, 
directly applying pinhole models to wide-FoV images often results in downgraded performance due to object deformations and image distortions. Previous work~\cite{yang2019can, yang2019pass} propose the Panoramic Annular Semantic Segmentation (PASS) framework using a panoramic annular lens system. ~\cite{jaus2021panoramic} introduces panoramic panoptic segmentation for panoramic views. In~\cite{zhang2022bending, zhang2024behind}, deformable modules are proposed to enhance the distortion perception capability in transformers. Zheng \textit{et al.}~\cite{zheng2025open} introduce the Open Panoramic Segmentation (OPS) task in an open-vocabulary context. 
Apart from panoramas, previous methods~\cite{kumar2021omnidet, deng2019restricted} revolve around multi-task on fisheye images, while a pretraining paradigm~\cite{paul2023semantic} and semi-supervised approaches~\cite{playout2021adaptable, paul2024fishsegssl} are introduced to address the challenges of fisheye data. 
Manzoor \textit{et al.}~\cite{manzoor2024deformable} substitutes the conventional convolutional modules in U-Net~\cite{ronneberger2015u} and residual U-Net~\cite{quan2021fusionnet} with restricted deformable convolution (RDC)~\cite{deng2019restricted}, imparting fisheye-distortion awareness to the model. Departing from these approaches, we are the first to propose a unified model applicable to all wide-FoV inputs, including panoramic and fisheye images.

\subsection{Vision Mambas} \label{related:mamba}
CNN~\cite{krizhevsky2012imagenet} have achieved notable success in computer vision with efficient local feature extraction and hierarchical learning, while Transformer~\cite{vaswani2017attention} excel at modeling long-range dependencies despite their quadratic complexity. Recently, the state-space-model (SSM)-based Mamba~\cite{gu2023mamba} architecture has emerged as a promising alternative, offering linear complexity and parallel computation through its selective scan mechanism, enabling efficient long-sequence modeling. The application of Mamba in sequence modeling has naturally extended to vision tasks. The serialized processing of input tokens in the Mamba architecture presents notable challenges when transferring to vision tasks. Zhu \textit{et al.}~\cite{vim} serialize the image using patch embedding~\cite{dosovitskiy2020image}, then scan the sequence from two directions. In~\cite{liu2024vmamba}, a quadri-directional scanning mechanism is proposed to supersede Mamba's initial 1D scan operation to 2D scan. Subsequent work~\cite{huang2024localmamba, pei2024efficientvmamba, hu2024zigma, li2024videomamba, zhang2025motion, yang2024vivim, chen2024rsmamba} introduce various scanning methods to explore variations of Mamba in visual representations. However, these works have primarily leveraged the efficiency of the Mamba architecture from an encoder perspective, while its application in the decoder remains relatively constrained. Besides, while benefiting from the efficiency of the Mamba architecture, its application in wide-FoV dense prediction tasks remains unexplored. Therefore, for the first time, we propose a mamba-based decoder to address wide-FoV image segmentation. 

%% file: sec/3_methodology.tex
\section{Methodology}
\label{sec:methodology}
\subsection{Overall Architecture} \label{sec3.1:overall}
An overview of our Deformable Mamba architecture is presented in Fig.~\ref{fig3}, which illustrates how we make Mamba-based models deformable and available for wide-FoV semantic segmentation. It mainly follows the Encoder-Decoder architecture. The backbone selection is flexible for different methods like CNN-, Transformer- or Mamba-based architectures. In this work, we explore with VMamba~\cite{liu2024vmamba} as the backbone and our proposed Deformable Mamba Decoder. In our framework, there are two model sizes: DMamba-M (\textbf{M}ini) and DMamba-T (\textbf{T}iny). 

Specifically, given the wide-FoV input image of shape $H \times W \times 3$, a stem module is firstly utilized to patchify the original image into the shape of $\frac{H}{4} \times \frac{W}{4} \times 3$, the encoder gradually down-samples feature maps in the $l^{th}$ stage with channel dimensions $C_l \in \{96, 192, 384, 768\}$ and resolutions $R_l\in \{\frac{H}{4} \times \frac{W}{4}, \frac{H}{8} \times \frac{W}{8}, \frac{H}{16} \times \frac{W}{16}, \frac{H}{32} \times \frac{H}{32} \}$. These features are hierarchically fused in our proposed Deformable Mamba Decoder. The output features from the decoder's final layer are first processed through a fully connected layer and subsequently transformed via feature channel convolutions to generate the predicted semantic map.

\subsection{Deformable Mamba Decoder}
\label{sec3.2}
The Deformable Mamba Decoder comprises four efficient fusion and upsampling modules. Within this framework, we propose a novel \textbf{Deformable Mamba Fusion (DMF)} block, which effectively integrates multi-scale features while enabling the decoder to adaptively process distorted data. Considering the diverse types and degrees of distortion present in different wide-FoV data, our method is not tailored to any specific dataset. Instead, we incorporate distortion awareness into the model in a simple, broadly applicable manner. 

As shown in lower part of Fig.~\ref{fig3}, the DMF block takes two inputs, $E_i$ and $D_j$, where $E_i$ represents the multi-scale feature maps output by the encoder, and $D_j$ denotes the features fused by the DMF block. Here, indices $i$ and $j$ range from 1 to 4 and 0 to 3, respectively, following the constraint $i + j = 4$. Specifically, for the first stage of DMF block, both inputs are derived from the feature maps of the last encoder layer, \textit{i.e.}, $E_4 = D_0$. To maximize efficiency by minimizing decoder computational complexity, we employ only a single quadri-directional scanning layer, originally introduced in the 2D selective scan (SS2D) block by Liu \textit{et al.}~\cite{liu2024vmamba} for embedding feature $D_j$ into the state space. 

\begin{figure}[t]
  \centering
  \includegraphics[width=\linewidth]{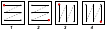}
  \caption{For an embedded 2D sequence, Mamba~\cite{gu2023mamba} employs a uni-directional scan from start to end (1).~\cite{vim} utilizes a bi-directional scan combining (1) and (2), while~\cite{liu2024vmamba} adopts a quadri-directional scanning incorporating (1), (2), (3), and (4). }
  \label{fig4}
\end{figure}

We posit that enabling model distortion-aware capability hinges on providing the model with sufficient deformable capacity, thereby endowing the model with the intrinsic ability to adaptively learn and generalize from distorted data representations. Deformable Convolutions (DCNs)~\cite{dai2017deformable, zhu2019deformable, wang2023internimage, xiong2024efficient} render the rigid kernels in conventional convolution more flexible and dynamically adaptive by introducing data-dependent offsets, which allows the fixed sampling locations to be dynamically adjusted based on the data. \cite{zhu2020deformable} provides the Transformer-based model with a data-driven receptive field, enabling the model's perception of image distortions and deformations. SSM, which describes the dynamics of a system, can be formulated as:
\begin{equation}
\begin{split}
    \mathbf{h}'(\mathbf{t}) &= \mathbf{A} \mathbf{h}(\mathbf{t}) + \mathbf{B} \mathbf{x}(\mathbf{t}), \\
    \mathbf{y}(\mathbf{t}) &= \mathbf{C} \mathbf{h}(\mathbf{t}), \\
\end{split}
\end{equation}
where $\mathbf{x}(\mathbf{t})$ and $\mathbf{y}(\mathbf{t})$ denote the inputs and outputs of the system, respectively, $\mathbf{h}(\mathbf{t})$ is the inner states of the system, $\mathbf{A}, \mathbf{B}, \mathbf{C}$ are the system parameters. 

Mamba~\cite{gu2023mamba} and its visual variants~\cite{vim, liu2024vmamba, huang2024localmamba, pei2024efficientvmamba, hu2024zigma, li2024videomamba, zhang2025motion, yang2024vivim, chen2024rsmamba} are based on SSM. As shown in Fig.~\ref{fig4}, while methods such as~\cite{vim, chen2024rsmamba} have attempted to model the spatial information of images by expanding scanning paths, the inherent one-by-one token processing makes it more challenging to integrate a learnable deformation mechanism into state-space modeling, compared to CNN- and Transformer-based architectures. 
To address this, we adopt a compromise solution by following~\cite{zhang2024behind, zhang2022bending, zheng2025open}, we incorporate deformable convolution to augment the model's deformable capacity. In particular, we feed the encoder feature $E_i$ into one single DCNv2~\cite{zhu2019deformable} layer with a kernel size of $\mathbf{3 \times 3}$, which can be mathematically formulated as follows:

\begin{equation}
    \mathbf{E_{out}}(\mathbf{p}) = \sum^K_{k=1}\mathbf{w}_k\mathbf{m}_k\mathbf{E_{in}}(\mathbf{p}+\mathbf{p}_k+\Delta\mathbf{p}_k),
\end{equation}
where $\mathbf{p}$ represents the current location in feature map, $\mathbf{w_k}$ and $\mathbf{p_k}$ denote the weight and initial offsets for the $\mathbf{k}$-th location, respectively. The parameter $\mathbf{m_k}$ is a learnable modulation scalar that modulates the input feature amplitudes from different spatial locations. In our implementation, $\mathbf{K}=9$ and $\mathbf{p_k} \in {(-1, -1), ..., (1, 1)}$ defines a $\mathbf{3 \times 3}$ convolutional kernel with dilation 1. 

The outputs from both branches, each with dimensions $\{h \times w \times c\}$, are concatenated to form a tensor with dimensions $\{h \times w \times 2c\}$, which is subsequently processed by an upsampling module. To fully preserve the feature information from both branches while endowing the model with additional learnable capacity, we refrain from employing parameter-free upsampling methods such as bilinear interporlation. Instead, after an initial feature fusion via a series of $\mathbf{3 \times 3}$ convolutions, we enhance the resolution of feature maps using the Pixelshuffle~\cite{shi2016real} operation, which can be formulated as:

\begin{equation}
    \mathcal{PS}(T)_{h, w, 2c} = T_{2h, 2w, \frac{c}{2}},
\end{equation}
where $\mathcal{PS}$ is an periodic shuffling operator that rearranges the elements of a input tensor $T$ from shape $\{h \times w \times 2c\}$ to $\{2h \times 2w \times \frac{c}{2}\}$.

The upsampled features are linearly projected as the output of the DMF block. Uniquely, the feature map in the final stage, after undergoing convolutional fusion, is not subjected to the PixelShuffle operation for channel reduction and resolution enhancement. It is worth noting that in all the experiments presented in Sec.~\ref{sec:experiments}, the decoder architecture is kept consistent, with no dataset-specific modifications. The only adjustment made is to the final output layer, which is tailored to accommodate the varying semantic categories across the datasets.

Our Deformable Mamba Decoder leverages the advantages of the Mamba architecture’s linear computational complexity, resulting in superior computational efficiency. Simultaneously, it addresses the challenge that Mamba, in contrast to CNNs and Transformers, lacks the architectural ability to provide deformation awareness at the feature level. Besides, we introduce distortion awareness exclusively from the decoder perspective, decoupling it from the overall framework and thus enhancing the model's flexibility and adaptability. Comprehensive qualitative and quantitative analyses in Sec.~\ref{sec:experiments} showcase the exceptional performance and efficiency of our decoder.

%% file: sec/4_experiments.tex
\section{Experiment Results}
\label{sec:experiments}
To validate our deformable Mamba method, we compare with state-of-the-art approaches  on three 360{\textdegree} panoramic and two 180{\textdegree} fisheye datasets. Additionally, we conduct experiments using CNN-, Transformer-, and Mamba-based architectures (e.g., ResNet, Swin, VMamba) to validate the efficiency and effectiveness of our Deformable Mamba Decoder. 

\subsection{Datasets}

\noindent\textbf{Stanford2D3D}~\cite{armeni2017joint} consists of 1,413 indoor 360{\textdegree} images with a resolution of $512{\times}1024$, encompassing 13 distinct object categories. 

\noindent\textbf{Matterport3D}~\cite{chang2017matterport3d} has 10,615 indoor 360{\textdegree} images. Following~\cite{teng2024360bev, guttikonda2024single}, 7,829 of panoramas with a resolution of 1024×2048 are used for training and 772 for evaluation.

\noindent\textbf{SynPASS}~\cite{zhang2024behind} is a synthetic outdoor 360{\textdegree} dataset from CARLA~\cite{dosovitskiy2017carla}. It contains 9,080 images with a resolution of $1024{\times}2048$, covering various weather conditions (cloudy, foggy, rainy, sunny) and illumination (daytime, nighttime). 

\noindent\textbf{WoodScape}~\cite{yogamani2019woodscape} consists of 10K 180{\textdegree} images captured by four surround view cameras. The semantic labels have 9 categories. Since the test split is not available, 8,234 released images are reallocated into training and evaluation sets with an 80\% to 20\% split.

\noindent\textbf{SynWoodScape}~\cite{sekkat2022synwoodscape} is a synthetic version of the surround-view dataset. The released v0.1.1 version contains 
2,000 180{\textdegree} surround-view fisheye RGB images. We split them into training and validation sets with an 80\% to 20 \% ratio.

\subsection{Implementation Details} 
In this work, our models are trained and validated on three 360{\textdegree} datasets and two 180{\textdegree} datasets, covering both indoor and outdoor scenarios, as well as real and synthetic environments, using 4 A100 GPUs.
For all experiments, we employed cross-entropy as the loss function. To validate the effectiveness of our framework, we deliberately avoided using any auxiliary losses. The models were optimized using AdamW with an initial learning rate of $6e^{-5}$ and weight decay of 0.01.
For Stanford2D3D and SynWoodscape, we trained for 80K iterations with a batch size of 2 per GPU. For Matterport3D, SynPASS, and Woodscape, we extended the training to 160K iterations, maintaining the same batch size of 2 per GPU. The learning rate schedule consisted of a linear warm-up for the first 1.5K iterations, followed by a polynomial decay with a power of 0.9.

\subsection{Effectiveness and Robustness analysis}  
\label{subsec: Effectiveness}

\input{tables/s2d3d_resnet50}
\input{tables/s2d3d_swin}

\input{tables/s2d3d_vmamba}

\noindent \textbf{Adaptability to different backbones.}
As mentioned in Sec.~\ref{sec:intro} and Sec.~\ref{sec:methodology}, decoupling the deformable setting from the overall framework and introducing it exclusively from the decoder perspective somewhat reduces the model's capability to perceive distortions in wide-FoV data. The advantage, however, lies in enhancing the model's flexibility. Through the plug-in distortion-aware decoder, we enable the model to adaptively perceive image deformations for different backbones. On the Stanford2D3D panoramic dataset, we compared our Deformable Mamaba Decoder with UperHead~\cite{xiao2018unified}, MusterHead~\cite{xu2022uperformer}, CGRHead~\cite{ni2024context}, evaluating their adaptability to CNN-, Transformer-, and Mamba-based backbones. As shown in Tables~\ref{tab:resnet}, \ref{tab:swin}, and \ref{tab:vmamba}, compared to decoders lacking distortion perception, our method achieves the best performance across ResNet-50~\cite{he2016deep}, Swin-T~\cite{liu2021swin}, and VMamba-T~\cite{liu2024vmamba} backbones, which demonstrates the significance of distortion perception for wide-FoV image understanding.

\noindent \textbf{Efficiency of the decoder.}
We emphasize that an efficient decoder is crucial for alleviating the overall model complexity. For instance, Liu \textit{et al.} ultilize UperHead as the decoder to evaluate VMamba's application in semantic segmentation tasks. As shown in Table~\ref{tab:vmamba}, with VMamba as the backbone, the model has only 25.3G FLOPs, while the decoder UperHead, with 206.9G FLOPs, becomes the bottleneck of the models' computational complexity. While MusterHead and CGRHead reduce computational complexity, they exhibit strong sensitivity to the model scale. As evidenced in Table~\ref{tab:resnet}, their parameter count experiences substantial growth with increasing embedding dimensions. In contrast, our proposed decoder showcases superior compatibility with diverse model scales while retaining lower temporal and spatial complexity.

Based on the results, our method provides substantial distortion-aware capability for CNN-, Transformer-, and Mamba-based models while maintaining excellent efficiency, demonstrating its adaptability and robustness.

\subsection{Qualitative Results}
\label{subsec:qualitative_results}

\begin{figure}[h]
  \centering
  \includegraphics[width=\linewidth]{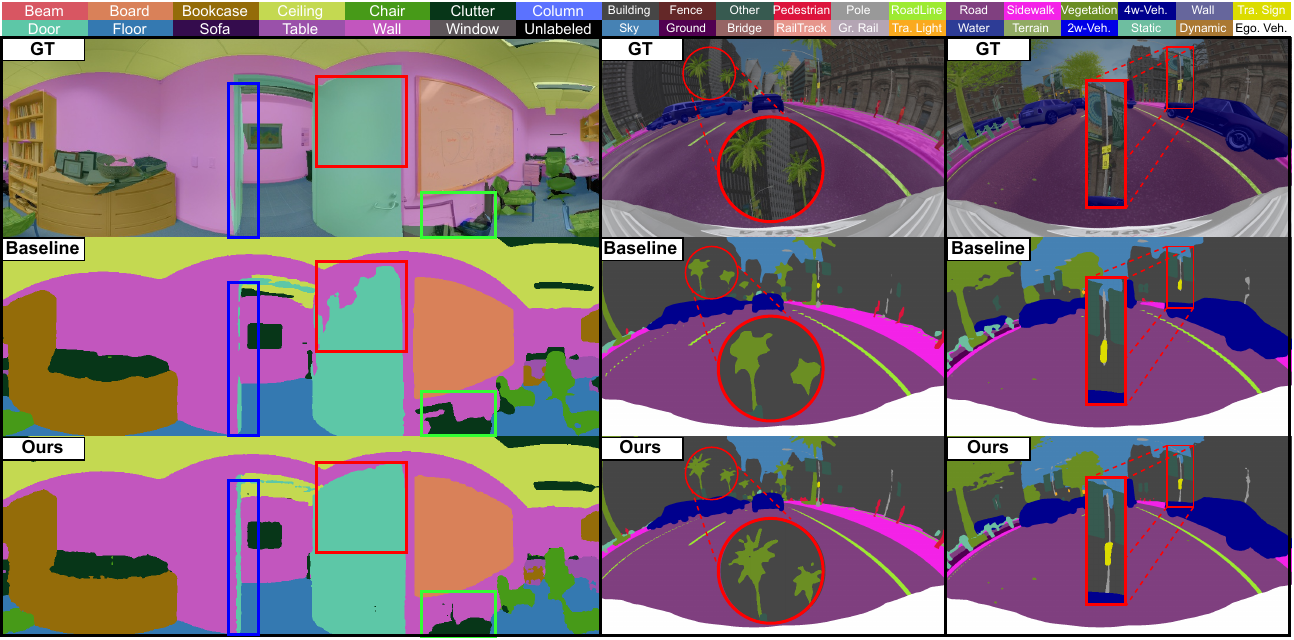}
  \vskip -0.5em
  \caption{Visualization of the wide-FoV segmentation results. From left to right: one 360{\textdegree} and two 180{\textdegree} segmentation. }
  \label{fig_vis}
\end{figure}

\noindent \textbf{Visualization of distortion-aware capability}.
To showcase the distortion-aware capability of our decoder, we present a visual comparison against UperHead~\cite{xiao2018unified} lacking deformable desin on the 360{\textdegree} Stanford2D3D and 180{\textdegree} SynWoodScape datasets. As shown in Fig.~\ref{fig_vis}, the baseline model struggles to accurately segment narrow and elongated areas such as door frames, vegetation contours, traffic light poles, and lane markings. In contrast, our proposed DMamba with Deformable Mamba Decoder effectively addresses these challenges, significantly improving segmentation performance on small and elongated objects.

\begin{figure}[t]
    \centering
    \begin{subfigure}{\linewidth}
        \centering
        \includegraphics[width=\linewidth]{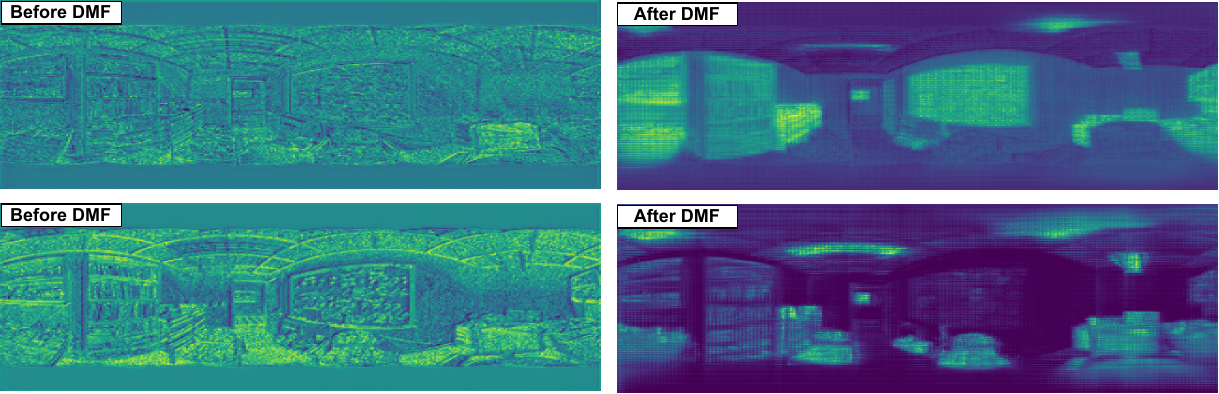}
        \caption{Visualization of feature maps in indoor scenarios}
        \label{fig6a}
    \end{subfigure}

    \begin{subfigure}{\linewidth}
        \centering
        \includegraphics[width=\linewidth]{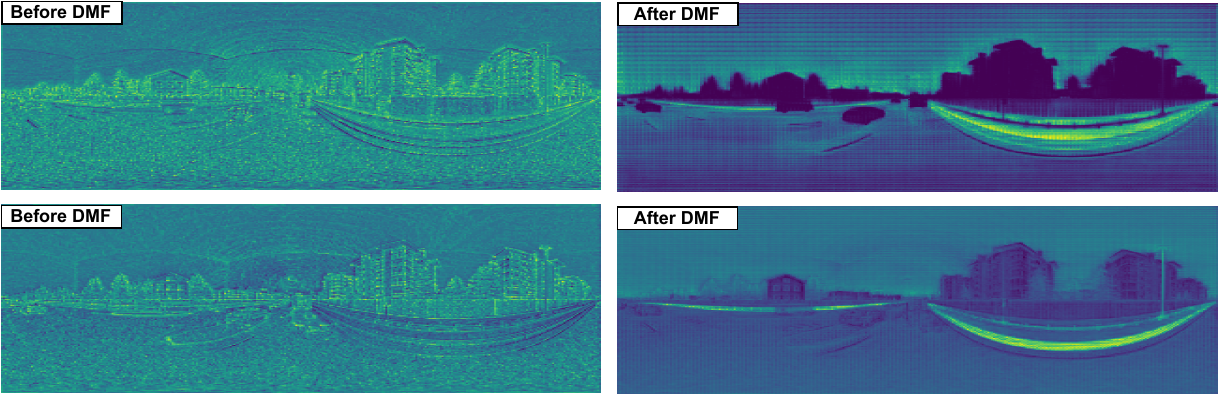}  
        \caption{Visualization of feature maps in outdoor scenarios}
        \label{fig6b}
    \end{subfigure}
    
    \caption{\textbf{Visualization of feature maps}. From left to right: features extracted by Mamba-based encoder (\textbf{Before DMF}) and features fused by the DMF module (\textbf{After DMF}). Effective detection and recognition of small objects in indoor scenarios (a), such as pictures on walls, requires the model to exhibit enhanced local perception capabilities. In contrast, outdoor scenarios (b) captured by wide-FoV cameras are often characterized by significant object deformation and distortion. To facilitate accurate scene understanding in such cases, the model must be equipped with a robust ability to handle and perceive distortion effectively.}
    \label{vis_fm}
\end{figure}

\noindent \textbf{Analysis of feature maps}. 
As shown in Figure~\ref{vis_fm}, to illustrate how the DMF module enhances distortion awareness, we visualize the channel map extracted from the encoder and the corresponding output features of DMF block. Specifically, we present the results of channels $\#6$ and $\#68$ in the second layer of the first decoder stage.
Prior to passing through the DMF block, the features primarily focus on global structures while lacking adequate emphasis on crucial regions such as small objects in cluttered indoor environments~\ref{fig6a} or severely distorted areas in wide-FoV outdoor scenes~\ref{fig6b}. Without explicit distortion-aware adaptation, the feature representation remains less effective in preserving fine-grained spatial details and compensating for geometric distortions. The model shows enhanced attention to objects at different scales after traversing the DMF block.

\subsection{Quantitative Results} \label{subsec: Quantitative}
We extensively evaluate our proposed method across the five aforementioned datasets, addressing the critical challenge of data heterogeneity prevalent in existing methods. Through meticulous comparative experiments with state-of-the-art approaches and a careful re-implementation of~\cite{guo2022segnext, zheng2024semantics}, we systematically investigate the performance variations induced by diverse data distributions and types, thereby providing comprehensive insights into model generalizability and robustness.

\input{tables/s2d3d_mp3d}

\noindent \textbf{Results on 360{\textdegree} Stanford2D3D dataset.} 
Table~\ref{tab:s2d3d_mp3d} shows superior performance of our method compared to both distortion-aware~\cite{eder2020tangent, sun2021hohonet, zheng2023complementary, zhang2022bending, zhang2024behind, teng2024360bev, li2023sgat4pass, zheng2024semantics} and non-distortion-aware~\cite{guo2022segnext, liu2024vmamba} methods. Specifically, our DMamba-M model with the VMamba-T as backbone achieves a significant mIoU improvement of ${+}2.9\%$ to ${+}13.7\%$ over panoramic baselines. 
Notably, compared to the VMamba~\cite{xiao2018unified}, our lightweighted decoder with the same backbone achieves a ${+}2.5\%$ higher mIoU while reducing overall parameters by 20\% and FLOPs by 86\%. Furthermore, our DMamba-T model, which has a comparable parameter count to VMamba, demonstrates even better performance with a ${+}3.4\%$ mIoU improvement. 

\noindent \textbf{Results on 360{\textdegree} Matterport3D dataset.} 
While our DMamba-M demonstrates remarkable performance on Stanford2D3D dataset, the challenge becomes more pronounced on large-scale Matterport3D panoramic datasets, where we observe a relatively modest improvement margin of ${+}0.3\%$ mIoU over the previous SOTA model SegNeXt as shown in Table~\ref{tab:s2d3d_mp3d}. Remarkably, our DMamba-T achieves more substantial gains of ${+}0.9\%$ and ${+}1.7\%$ mIoU compared to SegNeXt and VMamba-T respectively, with comparable model capacities.

\input{tables/synpass}
\input{tables/woodscape}
\input{tables/synwoodscape}

\noindent \textbf{Results on 360{\textdegree} SynPASS dataset.} 
Beyond real-world datasets, we evaluate the generalizability of our method on synthetic and outdoor scenes on the 360{\textdegree} SynPASS dataset. It is worth noting that despite SegNeXt's superior performance over 360SFUDA in real-world scenarios, it exhibits a substantial performance degradation on the synthetic dataset. This discrepancy underscores the sensitivity of certain architectures to data distribution shifts. In contrast, our method, as shown in Table~\ref{tab:synpass}, achieves the state-of-the-art mIoU of 42.78\%, surpassing all existing panoramic- and pinhole-based methods. Explicitly, our DMamba-T overall achieves top scores on \textit{pedestrian, pole, traffic sign, static} and \textit{dynamic}. For \textit{traffic sign} with slender properties, our model enjoys more than ${+}6.0\%$ mIoU.

\noindent \textbf{Results on 180{\textdegree} WoodScape dataset.} 
In Table~\ref{tab:woodscape}, we present a comparison of our method with other models using 180{\textdegree} fisheye inputs from the WoodScape dataset~\cite{yogamani2019woodscape}. Semi-supervised methods such as MeanTeacher~\cite{tarvainen2017mean}, CPS~\cite{chen2021semi}, and FishSegSSL~\cite{paul2024fishsegssl} show lower performance due to the limited data. Compared to multi-task models, our proposed DMamba, designed with deformable settings, outperforms all variants of the OmniDet~\cite{kumar2021omnidet} model. 
Although SegNeXt and 360SFUDA perform well on the WoodScape dataset, our method, with a more compact architecture still achieves a higher mIoU by ${+}0.9\%$ and ${+}0.8\%$, respectively. The DMamba-T variant further advances the state-of-the-art mIoU to 83.9\%, confirming DMamba's effectiveness in capturing heavily distorted object information in wide-FoV segmentation.

\noindent \textbf{Results on 180{\textdegree} SynWoodScape dataset.} 
Due to the lack of baseline comparisons on SynWoodScape, we limited our evaluation to SegNeXt and 360SFUDA. As shown in Table~\ref{tab:synwoodscape}, similar to the results on SynPASS, SegNeXt and 360SFUDA perform similarly on real 180{\textdegree} datasets, while the former exhibits a significant performance degradation on synthetic data. This further confirms the challenge of achieving consistently high performance when data distribution shifts. In comparison, our DMamba-M of similar scale outperforms them by ${+}4.36\%$ and ${+}1.73\%$ mIoU, respectively, achieving an improved segmentation accuracy of 69.20 mIoU on SynWoodScape.

These consistent improvements across all five wide-FoV datasets demonstrate the effectiveness of our proposed framework in unifying wide-FoV segmentation.

\subsection{Ablation Study}
To further analyze the effect of different components on our proposed framework, we conduct ablation study on the Stanford2D3D dataset. 

\noindent \textbf{Analysis of scanning mechanism.}
To ensure consistency, we adopt the same scanning method as in~\cite{liu2024vmamba}, as our model primarily uses its Mamba-based encoder. As mentioned in Sec.~\ref{sec:intro}, the SSM-based Mamba processes input tokens sequentially, which is suitable for textual data. However, for image data, where tokens have spatial relationships, multi-directional scanning is more appropriate. To highlight the difference between various scanning methods on our framework, we conducted ablation experiments on the Stanford2D3D dataset. As shown in Table~\ref{tab:ablation_scan}, both uni-directional scanning and bi-directional scanning results in slight performance drop. 

\input{tables/ablation_scans}

\noindent \textbf{Study of defomable convolution.}
In Sec.~\ref{sec:intro} and Sec.~\ref{sec:related_work}, we outlined the limitations of tightly coupling deformable designs with the entire model and the challenges faced by Mamba-based models in integrating deformable mechanisms due to their sequential processing of input tokens. We propose to introduce sufficient distortion awareness into the model by paralleling DCN in the decoder and integrating it through feature fusion. We replace deformable convolution with traditional one as shown in Table~\ref{tab:ablation_dmf}, the results prove the effectiveness of this trade-off solution.

\input{tables/ablation_dmf}

\noindent \textbf{Analysis of upsampling methods.}
Sampling methods play a critical role in our approach. As mentioned earlier, our decoder utilizes two branches to achieve complete feature modeling and the incorporation of distortion awareness respectively. To optimize computational efficiency, we adopt a single-layer module in each branch, which reduces model complexity while limiting the model’s learnable capacity. As a result, we disregard parameter-free interpolation-based upsampling methods, opting instead to fuse features from both branches and employ periodic arrangement to preserve all information. In Table~\ref{tab:ablation_upsample}, we compare bilinear and bicubic methods, showing the importance of our design.
\input{tables/ablation_upsample}

%% file: tables/s2d3d_resnet50.tex
\begin{table}[tb!]
\caption{\textbf{Results on Stanford2D3D}, using \textbf{ResNet-50}~\cite{he2016deep} as the backbone with 23.5M parameters, 22.9G FLOPs and hierarchical embedding dimensions of \{256, 512, 1024, 2048\}. 
\vspace{-2mm}
}
\label{tab:resnet}
\centering
\renewcommand{\arraystretch}{1.0}
\resizebox{\linewidth}{!}{
\setlength{\tabcolsep}{6.0pt}
\begin{tabular}{l|c|c|cc}
\Xhline{1.2pt}
\multirow{2}{*}[-\baselineskip]{\textbf{Decoder}} & \multirow{2}{*}[-\baselineskip]{\textbf{Params (M)~${\downarrow}$}} & \multirow{2}{*}[-\baselineskip]{\textbf{FLOPs (G)~${\downarrow}$}} & \multicolumn{2}{c}{\textbf{Results}~${\uparrow}$} \\  
\cline{4-5}
 &  &  & \textbf{mIoU} & \textbf{Acc} \\  
\hline
UperHead~\cite{xiao2018unified} {\fontsize{8pt}{10pt}\selectfont (CVPR20)} & \textbf{40.5} & 250.7 & 53.34 & 62.42 \\
MusterHead~\cite{xu2022uperformer} {\fontsize{8pt}{10pt}\selectfont (CVPR21)} & 203.1 & 211.5 & 54.39 & 63.27 \\
CGRHead~\cite{ni2024context} {\fontsize{8pt}{10pt}\selectfont (CVPR23)} & 282.6 & \textbf{31.4} & 54.12 & 62.94 \\
\hline
\rowcolor{gray!20} \textbf{Ours} & 77.3 & 38.8 & \textbf{57.22} & \textbf{66.32} \\
\Xhline{1.2pt}
\end{tabular}}
\end{table}

%% file: tables/s2d3d_swin.tex
\begin{table}[tb!]
\caption{\textbf{Results on Stanford2D3D}, using \textbf{Swin-T}~\cite{liu2021swin} as the backbone with 27.5M parameters, 25.6G FLOPs and hierarchical embedding dimensions of \{96, 192, 384, 768\}. 
\vspace{-2mm}
}
\label{tab:swin}
\centering
\renewcommand{\arraystretch}{1.0}
\resizebox{\linewidth}{!}{
\setlength{\tabcolsep}{6.0pt}
\begin{tabular}{l|c|c|cc}
\Xhline{1.2pt}
\multirow{2}{*}[-\baselineskip]{\textbf{Decoder}} & \multirow{2}{*}[-\baselineskip]{\textbf{Params (M)~${\downarrow}$}} & \multirow{2}{*}[-\baselineskip]{\textbf{FLOPs (G)~${\downarrow}$}} & \multicolumn{2}{c}{\textbf{Results}~${\uparrow}$} \\  
\cline{4-5}
 &  &  & \textbf{mIoU} & \textbf{Acc} \\  
\hline
UperHead~\cite{xiao2018unified} {\fontsize{8pt}{10pt}\selectfont (CVPR20)} & 31.5 & 206.9 & 53.82 & 63.06 \\
MusterHead~\cite{xu2022uperformer} {\fontsize{8pt}{10pt}\selectfont (CVPR21)} & 19.1 & 21.4 & 52.36 & 54.54 \\
CGRHead~\cite{ni2024context} {\fontsize{8pt}{10pt}\selectfont (CVPR23)} & 40.6 & \textbf{5.0} & 54.26 & 62.90 \\
\hline
\rowcolor{gray!20} \textbf{Ours} & \textbf{11.2} & 6.0 & \textbf{55.09} & \textbf{65.30} \\
\Xhline{1.2pt}
\end{tabular}}
\end{table}

%% file: tables/s2d3d_vmamba.tex
\begin{table}[tb!]
\caption{\textbf{Results on Stanford2D3D}, using \textbf{VMamba-T}~\cite{liu2024vmamba} as the backbone with 29.5M parameters, 25.3G FLOPs and hierarchical embedding dimensions of \{96, 192, 384, 768\}. 
\vspace{-2mm}
}
\label{tab:vmamba}
\centering
\renewcommand{\arraystretch}{1.0}
\resizebox{\linewidth}{!}{
\setlength{\tabcolsep}{6.0pt}
\begin{tabular}{l|c|c|cc}
\Xhline{1.2pt}
\multirow{2}{*}[-\baselineskip]{\textbf{Decoder}} & \multirow{2}{*}[-\baselineskip]{\textbf{Params (M)~${\downarrow}$}} & \multirow{2}{*}[-\baselineskip]{\textbf{FLOPs (G)~${\downarrow}$}} & \multicolumn{2}{c}{\textbf{Results}~${\uparrow}$} \\  
\cline{4-5}
 &  &  & \textbf{mIoU} & \textbf{Acc} \\  
\hline
UperHead~\cite{xiao2018unified} {\fontsize{8pt}{10pt}\selectfont (CVPR20)} & 31.5 & 206.9 & 56.80 & 65.01 \\
MusterHead~\cite{xu2022uperformer} {\fontsize{8pt}{10pt}\selectfont (CVPR21)} & 19.1 & 21.4 & 58.12 & 66.97 \\
CGRHead~\cite{ni2024context} {\fontsize{8pt}{10pt}\selectfont (CVPR23)} & 40.6 & \textbf{5.0} & 56.45 & 65.07 \\
\hline
\rowcolor{gray!20} \textbf{Ours} & \textbf{11.2} & 6.0 & \textbf{59.30} & \textbf{68.99} \\
\Xhline{1.2pt}
\end{tabular}}
\end{table}

%% file: tables/s2d3d_mp3d.tex
\begin{table}[tb!]
\caption{
\textbf{Results on 360{\textdegree} segmentation datasets} of Stanford2D3D (\textbf{S2D3D}) fold-1 and 360FV-Matterport (\textbf{MP3D}). $^\dag$denotes our re-implementation. 
\vspace{-2mm}
}
\label{tab:s2d3d_mp3d}
\centering
\renewcommand{\arraystretch}{1.0}
\resizebox{\linewidth}{!}{
\setlength{\tabcolsep}{6.0pt}
\begin{tabular}{l|c|cc}
\Xhline{1.1pt}
\multirow{2}{*}[-\baselineskip]{\textbf{Method}} & \multirow{2}{*}[-\baselineskip]{\textbf{Backbone}} & \multicolumn{2}{c}{\textbf{mIoU}} \\ 
\cline{3-4}
 &  & \scalebox{0.8}{\textbf{S2D3D}} &  \scalebox{0.8}{\textbf{MP3D}} \\ 

\hline
Tangent~\cite{eder2020tangent} {\fontsize{8pt}{10pt}\selectfont (CVPR20)} & ResNet-101 & 45.6 & - \\
HoHoNet~\cite{sun2021hohonet} {\fontsize{8pt}{10pt}\selectfont (CVPR21)} & ResNet-101 & 52.0 & 44.1 \\
CBFC~\cite{zheng2023complementary} {\fontsize{8pt}{10pt}\selectfont (CVPR23)} & ResNet-101 & 52.2 & - \\
Trans4PASS~\cite{zhang2022bending} {\fontsize{8pt}{10pt}\selectfont (CVPR22)} & Trans4PASS & 52.1 & 41.9 \\
Trans4PASS+~\cite{zhang2024behind} {\fontsize{8pt}{10pt}\selectfont (PAMI24)} & Trans4PASS+ & 54.0 & 42.6 \\
SegFormer~\cite{xie2021segformer} {\fontsize{8pt}{10pt}\selectfont (NeurIPS21)} & MiT-B2 & 51.9 & 45.5 \\
360BEV~\cite{teng2024360bev} {\fontsize{8pt}{10pt}\selectfont (WACV24)} & MiT-B2 & 54.3 & 46.3 \\
SGAT4PASS~\cite{li2023sgat4pass} {\fontsize{8pt}{10pt}\selectfont (IJCAI23)} & MiT-B2 & 56.4 & - \\

360SFUDA$^\dag$~\cite{zheng2024semantics} {\fontsize{8pt}{10pt}\selectfont (CVPR24)} & MixT-B2 & 54.9 & 46.7 \\

360SFUDA$^\dag$~\cite{zheng2024semantics} {\fontsize{8pt}{10pt}\selectfont (CVPR24)} & MixT-B3 & 56.9 & 47.9 \\

 SegNeXt$^\dag$~\cite{guo2022segnext} {\fontsize{8pt}{10pt}\selectfont (NeurIPS22)} & MSCAN-B & 57.0 & 48.4 \\

VMamba$^\dag$~\cite{liu2024vmamba} {\fontsize{8pt}{10pt}\selectfont (NeurIPS24)} & VMamba-T & 56.8 & 47.6 \\

\hline
\rowcolor{gray!8} \textbf{DMamba-M (Ours)} & VMamba-T & \textbf{59.3} & \textbf{48.7} \\
\rowcolor{gray!8} \textbf{DMamba-T (Ours)} & VMamba-S & \textbf{60.2} & \textbf{49.3} \\
\Xhline{1.1pt}
\end{tabular}}
\end{table}

%% file: tables/synpass.tex
\begin{table*}[t]
\caption{
\textbf{Results on 360{\textdegree} segmentation dataset} SynPASS.  $^\dag$denotes our re-implementation.
\vspace{-2mm}
}
\label{tab:synpass}
\centering
\renewcommand{\arraystretch}{1.3}
\resizebox{\textwidth}{!}{
\setlength{\tabcolsep}{1mm}
\begin{tabular}{l|c|cccccccccccccccccccccc}
\Xhline{1.1pt}
\textbf{Method} & \textbf{\rotatebox{90}{mIoU}} &  \textbf{\rotatebox{90}{Building}} &  \textbf{\rotatebox{90}{Fense}} &  \textbf{\rotatebox{90}{Other}} &  \textbf{\rotatebox{90}{Pedestrian}} &  \textbf{\rotatebox{90}{Pole}} &  \textbf{\rotatebox{90}{RoadLine}}  &  \textbf{\rotatebox{90}{Road}} &  \textbf{\rotatebox{90}{SideWalk}} &  \textbf{\rotatebox{90}{Vegetation}} &  \textbf{\rotatebox{90}{Vehicles}} & \textbf{\rotatebox{90}{Wall}} &  \textbf{\rotatebox{90}{TrafficSign}} &  \textbf{\rotatebox{90}{Sky}} &  \textbf{\rotatebox{90}{Ground}} & \textbf{\rotatebox{90}{Bridge}} &  \textbf{\rotatebox{90}{RailTrack}} &  \textbf{\rotatebox{90}{GroundRail}} &  \textbf{\rotatebox{90}{TrafficLight}} &  \textbf{\rotatebox{90}{Static}} & \textbf{\rotatebox{90}{Dynamic}} &  \textbf{\rotatebox{90}{Water}} & \textbf{\rotatebox{90}{Terrain}} \\
\hline
PVT-T~\cite{wang2022pvt} {\fontsize{8pt}{10pt}\selectfont (CVM22)} & 32.37 & 74.83 & 19.94 & 00.24 & 21.82 & 13.15 & 62.59 & 93.14 & 49.09 & 67.27 & 46.44 & 01.69 & 09.63 & 96.09 & 00.18 & 02.64 & 08.81 & 61.11 & 14.09 & 12.04 & 00.99 & 05.05 & 51.33 \\

PVT-S~\cite{wang2022pvt} {\fontsize{8pt}{10pt}\selectfont (CVM22)} & 32.68	& 78.02	& 27.12	& \textbf{00.27} & 23.48 & 16.51 & 59.81 & 92.87 & 50.21 & 66.22 & 43.50 & 01.12 & 08.67 & 96.34 & 00.44 & 00.15 & 02.82 & 63.88 & 13.78 & 15.15 & 01.58 & 08.78 & 48.29 \\

SegFormer-B1~\cite{xie2021segformer} {\fontsize{8pt}{10pt}\selectfont (NeurIPS21)} & 37.36	& 78.24	& 20.59	& 00.00	& 38.28	& 21.09	& 68.72	& 94.50	& 59.72	& 68.43	& 67.51	& 00.83	& 09.86	& 96.08	& 00.56	& 01.38	& 20.79	& 69.59	& 23.38	& 19.91	& 01.38	& 08.97	& 52.07 \\

SegFormer-B2~\cite{xie2021segformer} {\fontsize{8pt}{10pt}\selectfont (NeurIPS21)} & 37.24	& 79.25	& 23.58	& 00.00	& 40.01	& 20.14	& 65.28	& 92.80	& 46.92	& 68.64	& 77.45	& 01.42	& 15.00	& 96.33	& 00.57	& 00.58	& 02.68	& 67.60	& 25.89	& 20.80	& 01.99	& 20.92 & 51.53 \\

Trans4PASS-T~\cite{zhang2022bending} {\fontsize{8pt}{10pt}\selectfont(CVPR22)} & 38.53	& 79.17	& 28.18 & 00.13 & 36.04 & 23.69 & 69.16 & 95.51 & 61.71 & 69.77 & 71.12 & 01.53 & 16.98 & 96.50 & 00.56 & 01.60 & 15.22 & 70.48 & 26.03 & 23.11 & 02.08 & 09.24 & 49.77 \\

Trans4PASS-S~\cite{zhang2022bending} {\fontsize{8pt}{10pt}\selectfont (CVPR22)} & 38.57	& 80.02	& 24.56	& 00.07	& 41.49	& 25.23	& 72.00	& 95.89	& 59.88	& 69.07	& 77.08	& 01.04	& 13.72	& 96.69	& 00.67	& 00.73	& 05.60	& \textbf{72.56} & 25.93 & 22.45 & 02.78 & 08.34 & 52.65 \\

Trans4PASS+(T)~\cite{zhang2024behind} {\fontsize{8pt}{10pt}\selectfont (PAMI24)} & 39.42	& 79.63	& 24.45	& 00.21	& 44.23	& 26.71	& 70.32	& 95.86	& 61.80	& 69.25	& 78.85	& 01.09	& 13.81	& 97.12	& 00.91	& \textbf{03.48}	& 19.32	& 72.44	& 21.08	& 25.56	& 02.67	& 05.03	& 53.20 \\

Trans4PASS+(S)~\cite{zhang2024behind} {\fontsize{8pt}{10pt}\selectfont (PAMI24)} & 40.72	& 80.91	& 20.78	& 00.23	& 45.36 & 24.08 & 72.51	& 96.79	& \textbf{67.15} & \textbf{70.46} & 81.39 & \textbf{04.28} & 26.19 & 97.21 & 01.24 & 01.74 & 16.56 & 67.08 & 28.64 & 23.68 & 03.35 & 08.48 & \textbf{57.57} \\

SegNeXt-B$^\dag$~\cite{guo2022segnext} {\fontsize{8pt}{10pt}\selectfont (NeurIPS22)} & 39.90 &82.62 &16.06 &00.12 &43.21 &26.15 &71.69 &97.10 &60.64 &67.56 &82.56 &00.32 &24.43 &97.23 &\textbf{02.03} &00.00 &\textbf{43.23} &47.90 &31.61 &26.50 &03.63 &10.56 &42.68
 \\

360SFUDA-B2$^\dag$~\cite{zheng2024semantics} {\fontsize{8pt}{10pt}\selectfont (CVPR24)} & 40.37 &82.69 &22.27 &00.16 &41.86 &28.19 &74.98 &97.13 &62.22 &70.05 &81.55 &04.21 &21.09 &97.45 &01.06 &00.49 &31.77 &51.08 &32.15 &24.88 &03.59 &10.02 &49.23 \\

360SFUDA-B3$^\dag$~\cite{zheng2024semantics} {\fontsize{8pt}{10pt}\selectfont (CVPR24)} & 41.33 &82.91 &\textbf{32.33} &00.23 &41.87 &27.21 &75.13 &97.37 &62.90 &69.34 &\textbf{82.95} &01.66 &26.03 &\textbf{97.56} &00.95 &00.10 &41.39 &64.02 &32.30 &26.74 &03.48 &16.76 &48.14 \\
\hline
\rowcolor{gray!8} DMamba-M(Ours) &  \textbf{41.72}	& \textbf{83.19} & 21.52 & 00.21 & \textbf{48.15} & \textbf{32.20} & \textbf{75.8} & \textbf{97.51} & 61.00 & 67.74 & 81.86 & 01.06 & \textbf{31.30} & 94.47 & 01.37 & 00.68 & 29.63 & 49.41 & \textbf{36.35} & \textbf{30.99} & \textbf{04.08} & 20.54 & 45.27 \\

\rowcolor{gray!8} DMamba-T(Ours) & \textbf{42.78} &\textbf{84.24} &22.57 &00.25 &\textbf{49.20} &\textbf{33.25} &\textbf{76.85} &\textbf{98.56} &62.05 &68.79 &82.91 &02.11 &\textbf{32.35} &\textbf{98.52} &\textbf{02.42} &01.73 &30.68 &50.46 &\textbf{37.40} &\textbf{32.04} &\textbf{05.85} &\textbf{21.59} &46.32

\\
\Xhline{1.1pt}
\end{tabular}}
\end{table*}

%% file: tables/woodscape.tex
\begin{table}[tb!]
\caption{
\textbf{Results on 180{\textdegree} segmentation dataset} WoodScape. 
$^\dag$denotes our re-implementation. 
\vspace{-2mm}
}
\label{tab:woodscape}
\centering
\renewcommand{\arraystretch}{1.2}
\resizebox{\columnwidth}{!}{
\setlength{\tabcolsep}{3mm}{
\begin{tabular}{l|c|c|c}
\Xhline{1.1pt}
\textbf{Method} & \textbf{Mutli-task} & \textbf{Supervised} & \textbf{mIoU} \\
\hline
MeanTeacher~\cite{tarvainen2017mean} {\fontsize{8pt}{10pt}\selectfont (NeurIPS17)} & \ding{55} & \ding{55} & 54.20 \\
CPS~\cite{chen2021semi} {\fontsize{8pt}{10pt}\selectfont (CVPR21)} & \ding{55} & \ding{55} & 60.31 \\
CPS with CutMix~\cite{chen2021semi} {\fontsize{8pt}{10pt}\selectfont (CVPR21)} & \ding{55} & \ding{55} & 62.47 \\
FishSegSSL~\cite{paul2024fishsegssl} {\fontsize{8pt}{10pt}\selectfont (J. Imaging24)} & \ding{55} & \ding{55} & 64.81 \\

OmniDet~\cite{kumar2021omnidet} {\fontsize{8pt}{10pt}\selectfont (RA-L21)} & \ding{55} & \ding{51} & 72.50 \\
\quad \textcolor{lightgray}{+ DTP~\cite{guo2018dynamic}} {\fontsize{8pt}{10pt}\selectfont (ECCV18)} & \textcolor{lightgray}{\ding{51}} & \textcolor{lightgray}{\ding{51}} & \textcolor{lightgray}{75.80} \\
\quad \textcolor{lightgray}{+ GradNorm~\cite{chen2018gradnorm}} {\fontsize{8pt}{10pt}\selectfont (PMLR18)} & \textcolor{lightgray}{\ding{51}} & \textcolor{lightgray}{\ding{51}} & \textcolor{lightgray}{75.90} \\
\quad \textcolor{lightgray}{+ Uncertainity~\cite{kendall2018multi}} {\fontsize{8pt}{10pt}\selectfont (CVPR18)} & \textcolor{lightgray}{\ding{51}} & \textcolor{lightgray}{\ding{51}} & \textcolor{lightgray}{76.10} \\
\quad \textcolor{lightgray}{+ VarNorm~\cite{kumar2021omnidet}} {\fontsize{8pt}{10pt}\selectfont (RA-L21)} & \textcolor{lightgray}{\ding{51}} & \textcolor{lightgray}{\ding{51}} & \textcolor{lightgray}{76.60} \\

SegNeXt-B$^\dag$~\cite{guo2022segnext} {\fontsize{8pt}{10pt}\selectfont (NeurIPS22)} & \ding{55} & \ding{51} & 82.32 \\

360SFUDA-B2$^\dag$~\cite{zheng2024semantics} {\fontsize{8pt}{10pt}\selectfont (CVPR24)} & \ding{55} & \ding{51} & 81.74 \\

360SFUDA-B3$^\dag$~\cite{zheng2024semantics} {\fontsize{8pt}{10pt}\selectfont (CVPR24)} & \ding{55} & \ding{51} & 82.43 \\

\hline
\rowcolor{gray!8} \textbf{DMamba-M (Ours)} & \ding{55} & \ding{51} & \textbf{83.21} \\
\rowcolor{gray!8} \textbf{DMamba-T (Ours)} & \ding{55} & \ding{51} & \textbf{83.90} \\
\Xhline{1.1pt}
\end{tabular}
}
}
\end{table}

%% file: tables/synwoodscape.tex
\begin{table}[t]
\caption{\textbf{Results on 180{\textdegree} segmentation dataset} SynWoodScape.
$^\dag$denotes our re-implementation. 
\vspace{-3mm}
}
\label{tab:synwoodscape}
\centering
\renewcommand{\arraystretch}{1.2}
\resizebox{\columnwidth}{!}{
\setlength{\tabcolsep}{12mm}{
\begin{tabular}{l|c}
\Xhline{1.1pt}
\textbf{Method} & \textbf{mIoU} \\
\hline
SegNeXt-B$^\dag$~\cite{guo2022segnext} {\fontsize{8pt}{10pt}\selectfont (NeurIPS22)} & 64.86 \\
360SFUDA-B2$^\dag$~\cite{zheng2024semantics} {\fontsize{8pt}{10pt}\selectfont (CVPR24)} & 66.56 \\
360SFUDA-B3$^\dag$~\cite{zheng2024semantics} {\fontsize{8pt}{10pt}\selectfont (CVPR24)} & 67.47 \\

\hline
\rowcolor{gray!8} \textbf{DMamba-M (Ours)} & \textbf{69.20} \\
\rowcolor{gray!8} \textbf{DMamba-T (Ours)} & \textbf{69.62} \\
\Xhline{1.1pt}
\end{tabular}
}
}
\end{table}

%% file: tables/ablation_scans.tex
\begin{table}[H]
\renewcommand{\arraystretch}{1.0}
\caption{
\textbf{Analysis of scanning methods}.
\vspace{-2mm}
}
\label{tab:ablation_scan}
\centering
\renewcommand{\arraystretch}{1.1}
\resizebox{0.98\columnwidth}{!}{%
\setlength{\tabcolsep}{6mm}{
\begin{tabular}{c|c|c}
\Xhline{1.1pt}
\textbf{Scan Method} & \textbf{mIoU} & \textbf{mAcc} \\
\hline
Uni-direction & 57.53 & 67.52\\
Bi-direction & 57.94 & 67.66\\
Quadri-direction (Ours) & \textbf{59.30} & \textbf{68.99} \\
\Xhline{1.1pt}
\end{tabular}
}
}
\end{table}

%% file: tables/ablation_dmf.tex
\begin{table}[H]
\caption{
\textbf{Ablation Study of deformable designs.}
\vspace{-2mm}
}
\label{tab:ablation_dmf}
\centering
\renewcommand{\arraystretch}{1.2}
\resizebox{0.98\columnwidth}{!}{
\setlength{\tabcolsep}{3mm}{
\begin{tabular}{c|c|c}
\Xhline{1.2pt}
\textbf{Deformable} & \textbf{mIoU} & \textbf{mAcc} \\

\hline
\ding{55} & 57.45 & 66.52\\
\ding{51} & \textbf{59.30} (\textcolor{red}{$+1.85\%$}) & \textbf{68.99} (\textcolor{red}{$+2.47\%$})\\

\Xhline{1.2pt}
\end{tabular}
}
}
\end{table}

%% file: tables/ablation_upsample.tex
\begin{table}[H]
\renewcommand{\arraystretch}{1.0}
\caption{
\textbf{Analysis of upsample methods}.
\vspace{-2mm}
}
\label{tab:ablation_upsample}
\centering
\renewcommand{\arraystretch}{1.2}
\resizebox{0.98\columnwidth}{!}{%
\setlength{\tabcolsep}{6mm}{
\begin{tabular}{c|c|c}
\Xhline{1.1pt}
\textbf{Upsample Method} & \textbf{mIoU} & \textbf{mAcc} \\
\hline
Bilinear Interpolation & 56.90 & 66.08\\
Bicubic Interpolation & 56.97 & 66.12\\
PixelShuffle (Ours) & \textbf{59.30} & \textbf{68.99} \\
\Xhline{1.1pt}
\end{tabular}
}
}
\end{table}

%% file: sec/5_conclusion.tex
\section{Conclusion}
\label{sec:conclusion}
In this work, we are the first to explore the unification of the wide field of view (FoV) segmentation tasks, including 180{\textdegree} and 360{\textdegree} image segmentation. Most existing models are tailored for pinhole camera images, making it challenging to extend their success to wide-FoV domains due to the severe image distortion and object deformation. We incorporate the model's distortion-aware capability exclusively from the decoder perspective, decoupling it from the overall architecture to enhance flexibility and facilitate seamless integration with various model architectures. Consistent improvements across five diverse wide-FoV datasets, ranging from indoor to outdoor, from synthetic to real-world scenes, demonstrate that our method effectively addresses the unique challenges of wide-FoV segmentation. 

\noindent\textbf{Limitations and Future Work.} 
Given the rapid advancement of Large Language Models(LLMs), integrating multi-modal LLMs presents a promising direction for future research. Their strong generalization capabilities and semantic understanding could significantly enhance our unified semantic segmentation framework, particularly in handling diverse wide-FoV segmentation tasks. This integration could bridge the gap between linguistic semantic understanding and geometric feature learning, potentially leading to more robust and versatile segmentation models.